\def\year{2022}\relax
\documentclass[letterpaper]{article} 
\usepackage{aaai22}  
\usepackage{times}  
\usepackage{helvet}  
\usepackage{courier}  
\usepackage[hyphens]{url}  
\usepackage{graphicx} 
\usepackage{natbib}  
\usepackage{caption} 
\DeclareCaptionStyle{ruled}{labelfont=normalfont,labelsep=colon,strut=off} 
\usepackage{algorithm}
\usepackage{algorithmic}
\usepackage{adjustbox}
\usepackage{multirow}

\usepackage{amssymb}
\usepackage{amsmath}
\usepackage{xcolor,colortbl}
\usepackage{verbatim}

\usepackage{newfloat}
\usepackage{listings}
\floatstyle{ruled}
\newfloat{listing}{tb}{lst}{}
\floatname{listing}{Listing}
\title{Improved Few-shot Segmentation by Redefinition of the Roles\\ of Multi-level CNN Features}

\author {
    Zhijie Wang\textsuperscript{\rm 1},
    Masanori Suganuma\textsuperscript{\rm 1,2},
    Takayuki Okatani\textsuperscript{\rm 1,2}
}
\affiliations {
    \textsuperscript{\rm 1} Graduate School of Information Sciences, Tohoku University \\
    \textsuperscript{\rm 2} RIKEN Center for AIP \\
    \{zhijie, suganuma, okatani\}@vision.is.tohoku.ac.jp
}

\usepackage{bibentry}

\begin{document}

\maketitle

\begin{abstract}
This study is concerned with few-shot segmentation, i.e., segmenting the region of an unseen object class in a query image, given support image(s) of its instances. The current methods rely on the pretrained CNN features of the support and query images. The key to good performance depends on the proper fusion of their mid-level and high-level features; the former contains shape-oriented information, while the latter has class-oriented information. Current state-of-the-art methods follow the approach of Tian et al., which gives the mid-level features the primary role and the high-level features the secondary role. In this paper, we reinterpret this widely employed approach by redefining the roles of the multi-level features; we swap the primary and secondary roles. Specifically, we regard that the current methods improve the initial estimate generated from the high-level features using the mid-level features. This reinterpretation suggests a new application of the current methods: to apply the same network multiple times to iteratively update the estimate of the object's region, starting from its initial estimate. Our experiments show that this method is effective and has updated the previous state-of-the-art on COCO-20$^i$ in the 1-shot and 5-shot settings and on PASCAL-5$^i$ in the 1-shot setting.
\end{abstract}

\begin{figure}[t]
\centering
\includegraphics[width=0.47\textwidth]{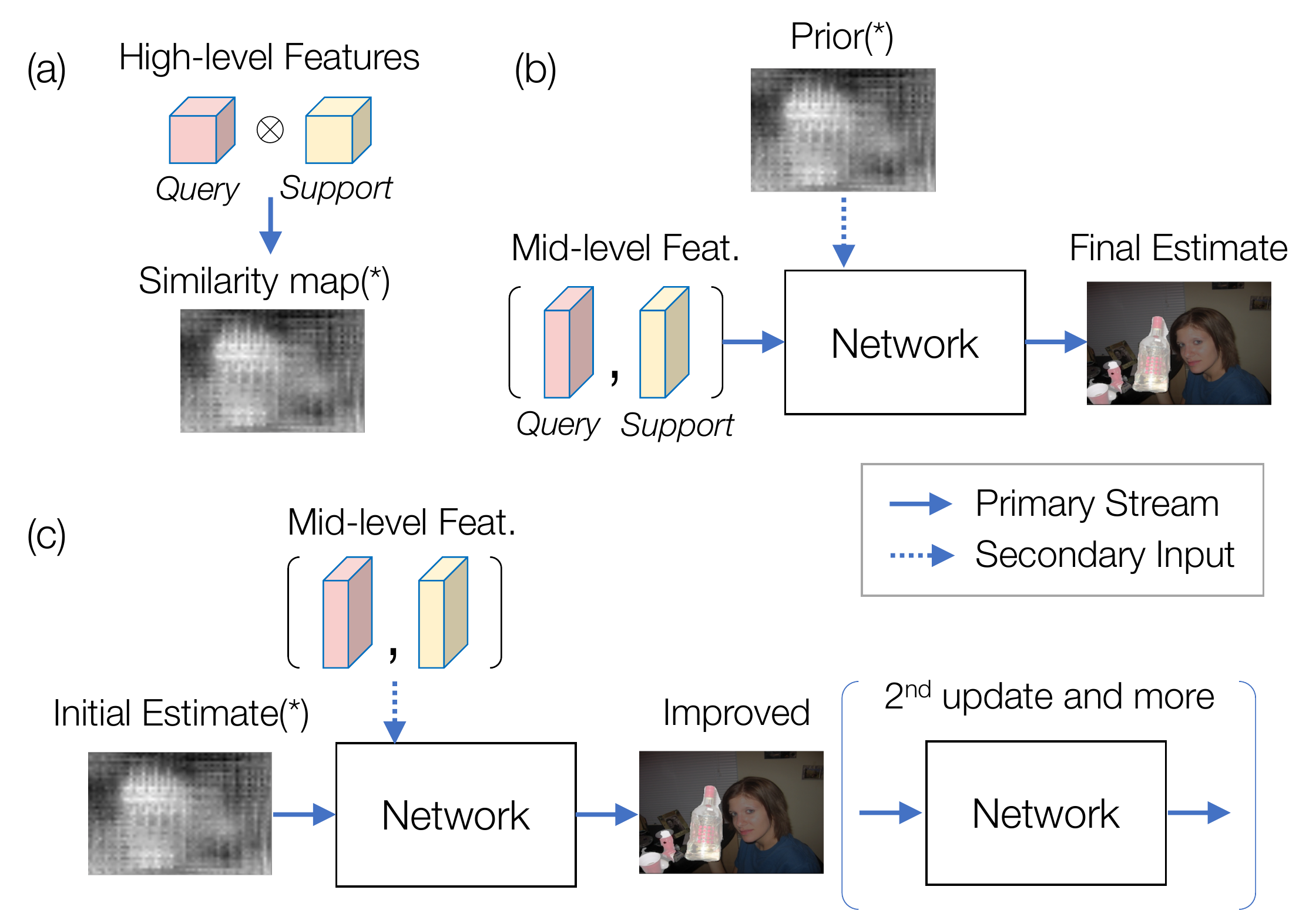}
\caption{
(a), (b) The conventional explanation of the current SOTA methods, i.e., PFENet etc. Given the support image(s), the target object region of the query image is estimated from their mid-level features, where the similarity map of their high-level features (i.e., ``prior'') serves as a prior. (c) Our interpretation: the same similarity map serves as an initial estimate, which is improved using the mid-level features. This enables iterative improvements. Note that the three entities with (*) are identical. }
\label{pic:introduction}
\end{figure}

\section{Introduction}

Semantic segmentation is one of the fundamental problems of computer vision, for which CNNs have been the main workhorse. The high-level features of CNNs convey information suitable for identifying object classes but lack detailed information about their shapes, whereas the opposite is true for their mid-level and lower features. CNN models designed and trained for semantic segmentation can properly utilize both the high-level and mid-level features, enabling accurate segmentation of objects.. 

In the problem of few-shot segmentation, we can use only a few labeled samples, called support images with segmentation masks, making a proper integration of the different level features more challenging. The comparison of the support and query images using only their mid-/low-level features or using only high-level features lead to inaccurate results. The key to success is how to fuse and use them adequately. 

Tian et al. propose a method named PFENet (Prior-guided Feature Enrichment Network) \cite{pfenet}, which has significantly raised the accuracy of few-shot segmentation, having established a new standard. Before PFENet, a good method to use the
high-level features was unknown, and their use reportedly leaded even to worse generalization. Therefore, researchers focused mainly on the mid-level features, although the high-level features should be useful for segmentation considering its nature of conveying more semantic information.

PFENet generates a single channel mask called a prior, which represents the similarity between the high-level features of the support and query images. It then fuses this prior mask with the mid-level features of the query and support images to create a new feature map, from which it predicts the object's region. Their fusion is conducted by the simple concatenation in the channel direction, followed by the application of a learnable module called feature enrichment module (FEM), where each of the fused components has an interaction with others. Most subsequent methods follow the same framework, such as ASGNet~\cite{asgnet}, SCL~\cite{scl}, SAGNN~\cite{sagnn}, etc. to name a few. 

However, {\em it is not very clear why this approach achieves such good performances.} There are two notable ingredients in their method. One is the way of utilizing the high-level features; instead of using them as they are, it uses their similarity map between the support and query images in all the subsequent computation. The other is how to fuse the similarity map with the mid-level features; in PFENet, they are simply concatenated in their channel dimension.

According to their paper \cite{pfenet}, the similarity map of the high-level features supposedly plays an auxiliary role, as it is called the prior and, moreover, PFENet is the abbreviation of ``prior-guided feature enrichment network.'' The authors arguably intend to use mid-level features primarily for segmentation, following earlier studies. Although it is not clearly stated so, they must suppose that the similarity map serves as the {\em prior for inferring the object's region} using the mid-level features. This explanation is adopted by many subsequent studies  \cite{asgnet,scl}.

In this paper, we reinterpret the approach as follows: {\em the similarity map of the high-level features plays the primary role, and the mid-level features play a secondary role.} Specifically, we regard the similarity map as an initial estimate of the object's region and consider PFENet\footnote{(We will consider only PFENet in this paper, but the discussions should be applicable to others.)} merely improves the estimate with the help of the mid-level features. By this new interpretation, we can obtain further improvements. Specifically, we propose a method that applies PFENet several times to iteratively update the estimate, starting from its initial estimate. We show that this method contributes to performance improvements without bells and whistles. There are several options with this method. One is how many times we iterate the inference. Another is whether we use different networks per each iteration or the same network at any iteration. The last is whether we replace the similarity map completely with the latest estimate of the object region or not. Our experiments show that any configuration improves the original PFENet; a particular configuration (i.e., one or two iterations, the same network, and the combined use of the original prior and the predicted mask) achieves the best performance.

\section{Related Work}

\subsection{Semantic Segmentation}
The task of semantic segmentation is to assign a class to each pixel in an image and CNN-based approaches have shown remarkable performance in recent years. Inspired by FCN~\cite{fcn}, which is the first work to employ a fully convolutional structure and achieves good performance, many methods have been designed based on FCN.
Recent mainstream is to introduce multi-scale feature aggregation~\cite{deeplab, pspnet,chen2018encoder,denseaspp,he2019adaptive,hrnet} and attention mechanisms~\cite{zhu2019asymmetric,expectation,danet, ocrnet} into a model.
While these methods have been making great progress, they heavily rely on large amounts of pixel-wise annotations to train the network. Moreover, these fully supervised methods generally show poor generalization to unseen categories that do not exist in the training set.

\subsection{Few-shot Segmentation}
Few-shot segmentation has been widely studied to deal with the above issues, where a network aims to segment foreground objects of an unseen class in a query image by using only a few labeled samples (i.e. support images).

OSLSM~\cite{bmvconeshot} is the pioneer work for few-shot segmentation and proposes a two-branched approach, in which the first branch takes support images to generate prototypes, and the second branch performs segmentation using them and a query image. PANet~\cite{panet} embeds different object classes into different prototypes with a pretrained encoder, and the query image will be labeled based on the distance between the representations of the query image and the prototypes. PFENet~\cite{pfenet} calculates the cosine similarity between high-level features extracted from support and query images to guide better segmentation results. To utilize more comprehensive information from the support images, SCL~\cite{scl} proposes a self-guided mechanism and produces new feature vectors for better segmentation. ASGNet~\cite{asgnet} proposes to condense the object features in support images into multiple feature vectors and allocate the most relevant features to each pixel in a query image, aiming at adaptively dealing with different object scales, shapes, and occlusions.

\section{Roles of Multi-level Features for Few-shot Segmentation}

\begin{figure*}[t]
\centering
\includegraphics[width=0.8\textwidth]{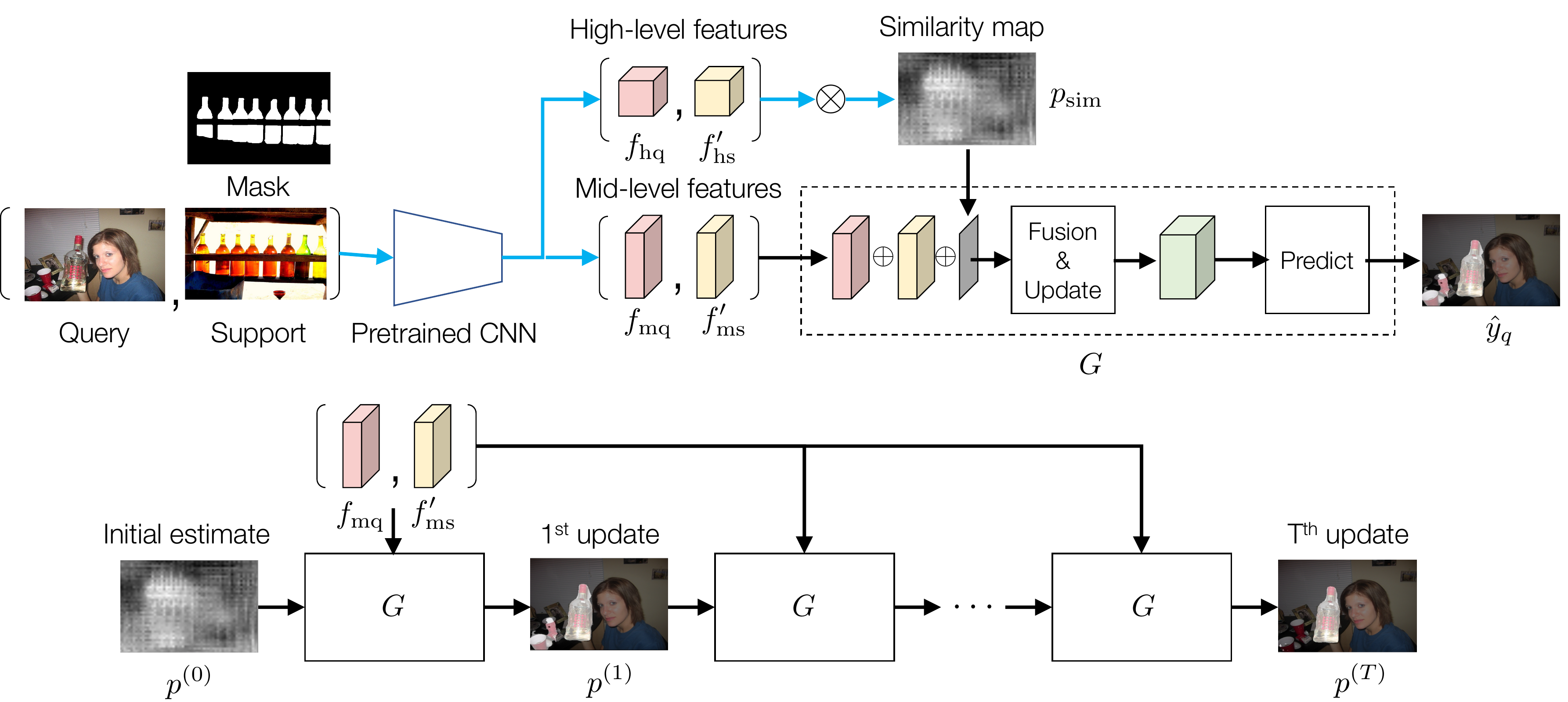} 
\caption{Upper row: The original pipeline of PFENet \cite{pfenet}. The left part with cyan-colored arrows shows the creation of inputs to the main body of the network $G$. Lower row: Our approach of using the network $G$ in a cascaded fashion to improve the initial estimate $p^{(0)}$, which is the same as $p_\mathrm{sim}$ in the upper row. }
\label{pic:overall}
\end{figure*}

\subsection{
Few-shot Segmentation}

Few-shot segmentation is the task of segmenting the region of a unseen object class in a query image, given $K$ support images of the same object instances. Specifically, we want to train a model on a training dataset $\mathcal{D}_\mathrm{train}$ and evaluate it on a dataset $\mathcal{D}_\mathrm{test}$, where their label sets $C_\mathrm{train}$ and $C_\mathrm{test}$ are disjoint from each other, i.e., $C_\mathrm{train} \cap C_\mathrm{test} = \varnothing$. The task is called 1-shot when $K=1$ and 5-shot when $K=5$.

The standard procedure for training \cite{bmvconeshot,pfenet} creates multiple training episodes by randomly choosing samples from $\mathcal{D}_\mathrm{train}$. It then trains a model using them. Each episode consists of a support set $S$ and a query set $Q$ for each class $c$; we usually consider the multiple classes in the training and the evaluation. The support set $S$ contains $K$ pairs of an image and its ground-truth mask, i.e., $S=\{(x^{(i)}_s, y^{(i)}_s)\}_{i=1}^K$, where $x^{(i)}_s$ and $y^{(i)}_s$ represent the $i$-th support image and its ground-truth mask, respectively.
The query set $Q$ contains an image and its ground-truth mask for the same class with $S$, i.e., $Q=\{(x_q, y_q)\}$.

We evaluate the performance of the trained model using $\mathcal{D}_\mathrm{test}$. We create multiple episodes from $\mathcal{D}_\mathrm{test}$ in the same way as the training episodes. For the sake of clarity, we denote $S_\mathrm{train}$ and $Q_\mathrm{train}$ to represent
the support set and the query set of a training episode, respectively, while we denote $S_\mathrm{test}$ and $Q_\mathrm{test}$ to represent those of a testing episode. We train a model using $S_\mathrm{train}$ and $Q_\mathrm{train}$ so that it will perform well on $S_\mathrm{test}$ and $Q_\mathrm{test}$ in a meta-learning fashion. Note that we cannot access to the ground-truth mask $y_q$ of any query image in $Q_\mathrm{test}$. 

\subsection{Revisiting the Prior-guided Feature Enrichment Network (PFENet)}

The PFENet is divided into two parts, i.e., the part for the feature extraction using a pretrained CNN and the subsequent part receiving the features and predicting the estimate $\hat{y}_q$ of the segmentation mask;
see the upper row of Fig.~\ref{pic:overall}. 

The second part is further divided into three parts, the generation of the prior, the fusion/update of the input features, and the prediction of the segmentation mask from the fused feature. We describe the first two below in more details. PFENet employs a plain stack of convolution layers for the third part.
We denote the combined network including the latter two (i.e., excluding the prior generation) by $G$ in what follows.

\subsubsection{Feature Extraction.}

In the feature extraction part, we input the query and the support images to the pretrained CNN to extract their the mid-level features $f_\mathrm{mq}$ and $f_\mathrm{ms}$ and the high-level features $f_\mathrm{hq}$ and $f_\mathrm{hs}$, respectively, as shown in Fig.~\ref{pic:overall}. Taking ResNet-50~\cite{resnet} as an example pretrained CNN, we use conv\_3x and conv4\_x for the middle-level features and conv5\_x for the high-level features. There is a segmentation mask for the target image. We neglect the background region of the support features by multiplying them with the mask as $f_{\mathrm{hs}}^{'} = f_{\mathrm{hs}} \odot y_{s}$ and $f_{\mathrm{ms}}^{'} = f_{\mathrm{ms}} \odot y_{s}$.

Tian et al. argue in \cite{pfenet} that the high-level features convey semantic information that is more class-specific than the middle-level features and that the high-level features will contribute more in identifying pixels of the target object class, whereas the mid-level features will contribute to the generalization on unseen classes. Then, PFENet uses the high-level features to calculate the prior $p_\mathrm{sim}$ for guiding the segmentation of its detailed shape using the mid-level features.



\subsubsection{Prior Generation.}

PFENet generates a similarity map between the the high-level features of the query and the support images. It first calculates the cosine similarity between any pixel pairs of the two features as
\begin{equation}\label{eq:cossim_cal_1}
\mathrm{sim}(f_{\mathrm{hq},i},f'_{\mathrm{hs},j}) = \frac{f_{\mathrm{hq},i}^\top f_{\mathrm{hs},j}^{'}}{{\lVert f_{\mathrm{hq},i} \rVert}{\lVert f_{\mathrm{hs},j}^{'} \rVert}}, 
\end{equation}
where $f_{\mathrm{hq},i}$ and $f'_{\mathrm{hs},j}$ $(\in \mathbb{R}^{\mbox{\tiny (num. of channels)}})$ indicate the $i$-th and $j$-th pixel features of $f_\mathrm{hq}$ and $f'_\mathrm{hs}$, respectivel; $i,j =1,\ldots,hw$. It then computes the maximum value over the support image for each location $i$ of the query image as 
\begin{equation}\label{eq:cossim_cal_2}
\mathrm{sim}_i = \max_{j\in\{1,\ldots,hw\}}\mathrm{sim}(f_{\mathrm{hq},i},f'_{\mathrm{hs},j})
\end{equation}
PFENet reshapes $[\mathrm{sim}_1,\ldots,\mathrm{sim}_{hw}]$ into $v_{\mathrm{sim}} \in \mathbb{R}^{h \times w}$ and scale it to the range of $[0,1]$ with min-max normalization as
\begin{equation}\label{eq:cossim_cal_3}
p_{\mathrm{sim}} = \frac{v_{\mathrm{sim}} - \mathrm{min}(v_{\mathrm{sim}})}{\mathrm{max}(v_{\mathrm{sim}}) - \mathrm{min}(v_{\mathrm{sim}}) + 1e^{-7}}.
\end{equation}




\subsubsection{Feature Fusion}

\begin{figure}[t]
\centering
\includegraphics[width=0.47\textwidth]{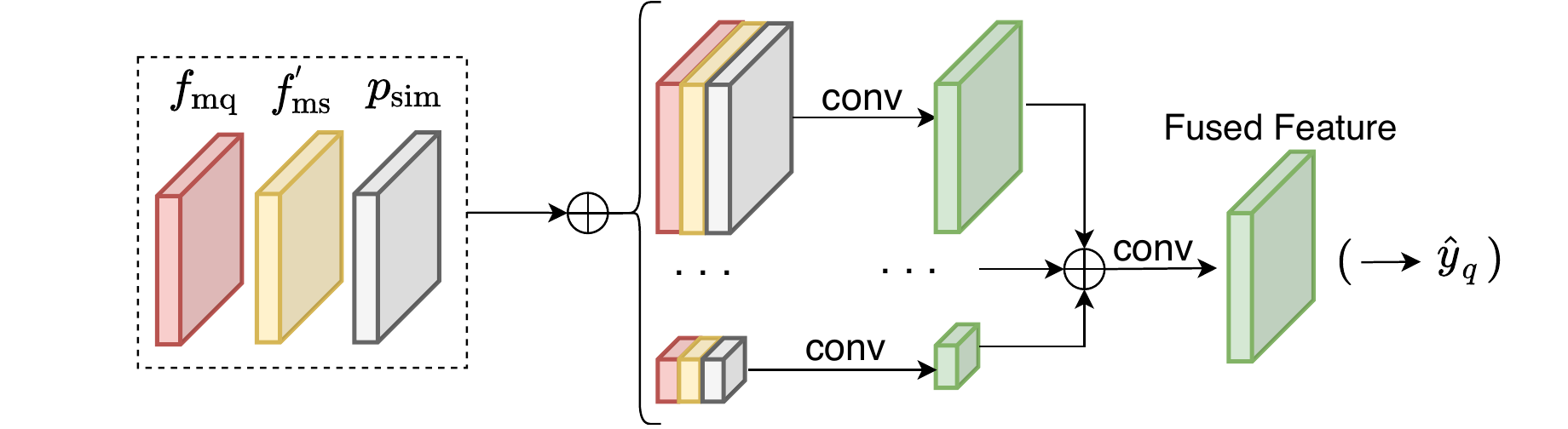} 
\caption{A simplified diagram of the feature enrichment module (FEM) employed in PFENet for the feature fusion. 
$\oplus$ means channel-wise concatenation.}
\label{pic:decoder}
\end{figure}

Receiving three inputs, the query mid-level feature $f_{\mathrm{mq}}$, masked support mid-level features $f_{\mathrm{ms}}^{'}$, and the prior $p_{\mathrm{sim}}$, PFENet fuses them with the module named feature enrichment module (FEM). FEM combines the pyramid pooling module \cite{pspnet} and the feature pyramid network \cite{fpn}, and performs multi-scale feature fusion. Fig.~\ref{pic:decoder} illustrates the structure of FEM. The three inputs are concatenated in their channel dimension as shown in the figure. FEM scales the concatenated feature map into multiple sizes = $\{60, 30, 15, 8\}$, and enriches it through the interactions between different channels and scales. See the original paper for more details.


\subsection{Reinterpreting PFENet} 

Toward the success of few-shot segmentation, a key question is how to fuse the mid-level and the high-level features extracted from the input images. We can make the following observations about how PFENet does this:
\begin{itemize}
\item It (more strictly, FEM of Fig.~\ref{pic:decoder}) receives three inputs, i.e., the mid-level features of the support and the query images and the similarity map of their high-level features. 
\item It fuses them by simple concatenation in their channel dimension to form a new feature map, which is then updated by a series of convolutions inside FEM.
\end{itemize}
We may state the intention behind the above treatments as follows: it uses the mid-level features primarily to predict the final segmentation mask while using the high-level features as a guide for it. In other words, {\em the mid-level features are the primary, and the high-level features are the secondary.} This design is also based on several earlier studies, such as the report of \cite{canet} that the direct use of the high-level features leads to inferior results.

Although the effectiveness of the approach is empirically well-confirmed, this explanation is not very persuasive. For example, what guidance is provided by the similarity map, and how is it effective on the inference based on mid-level features? Can we say what the ideal guidance is? These are not clear. All these seem to be left for the black box of FEM and the subsequent subnetwork. Let us remind that the high-level similarity map and the mid-level features are fused by simple channel-wise concatenation. They are fused on equal terms, and inherently there is no primary or secondary. 

Then, we propose here another interpretation of the approach. We regard {\em the similarity map play a leading role and the middle-level features are given a supporting role.}  Specifically, we regard the similarity map as a rough, initial estimate of the object's region, and PFENet improves its accuracy with the help of the mid-levels features. Indeed, the similarity maps (i.e., the priors) look like rough estimates of the object's regions with very obscure region boundaries, as shown in Fig.~\ref{pic:overall}.

Following this interpretation, we expect further improvements will be achievable by applying the method (i.e., PFENet) to the improved estimate again or even multiple times. The details are given below.


\color{black}

\subsection{Proposed Method} \label{sec_method}


\subsubsection{Base Method.}
The above interpretation of the framework of PFENet suggests that we perform the refinement of the initial estimate multiple times. Specifically, after we have run PFENet once, we input its output (i.e., the estimate of the object region) to PFENet and run it again. 
We may be able to repeat this process if necessary. 

The method is mathematically stated as follows. The network $G$ receives three inputs, i.e., two mid-level features, $f'_{\mathrm{ms}}$ (i.e., masked support), $f_\mathrm{mq}$ (query), and the latest estimate $p^{(t-1)}$ of the query object region, where $t(=1,\ldots,T)$ is the time step. It predicts the logits $z^{(t)}$ $\in \mathbb{R}^{2 \times h \times w}$ as
\begin{equation}\label{eq:dynamic_prior_1}
z^{(t)} = G(p^{(t-1)} \oplus f'_{\mathrm{ms}} \oplus f_{\mathrm{mq}}),
\end{equation}
then normalizes it as
\begin{equation}\label{eq:dynamic_prior_2}
p^{(t)} = \frac{\exp(z^{(t)}_1)}{
\exp({z}^{(t)}_0)+\exp(z^{(t)}_1)
},
\end{equation}
where $(z^{(t)}_0,z^{(t)}_1)$ are the binary elements at each pixel, and $p^{(t)}(\in \mathbb{R}^{1 \times h \times w})$ represents the probability that each pixel in the query image belongs to the target class.
Starting with $p^{(0)}\equiv p_\mathrm{sim}$, we iterate the update of $z^{(t)}$ by Eq. \ref{eq:dynamic_prior_1}.
We obtain a binary segmentation mask $\hat{y}_{\mathrm{q}}$ from $p^{(T)}$ as 
\begin{equation}\label{eq:result}
\hat{y}_{\mathrm{q}, i} =
\begin{cases}
    1 &\parbox{0.2\textwidth}{if $p^{(\mathrm{T})}_i > 0.5$,} \\
    0 &\parbox{0.2\textwidth}{otherwise},
\end{cases}
\end{equation}
where the subscript $i$ is the index of image pixels. 

This is the base version of our method. There are possible several extensions and options, which we will explain below. 


\subsubsection{How to Input the Latest Estimate.}

The underlying idea behind the repeated application of the network is that if more accurate input is given , the network can improve it to a further better estimate. 
However, the ideal input should not be the true segmentation map. (If it is, there is no need to use the mid-level features to improve it. The true map is unavailable, either.) Then, the question is how (in)accurate it should be. 

We consider a combination of $p_\mathrm{sim}$ and $p^{(t)}$ of Eq.~\ref{eq:dynamic_prior_2} and use it instead of $p^{(t)}$. 
Specifically, we use their element-wise multiplication 
\begin{equation}\label{eq:dynamic_prior_3}
p_\mathrm{aug}^{(t)} = (p^{(t)} \odot p_{\mathrm{sim}})_{\mathrm{norm}},
\end{equation}
where $p_\mathrm{aug} \in \mathbb{R}^{1 \times h \times w}$ and $\mathrm{norm}$ means the min-max normalization in Eq.~\ref{eq:cossim_cal_3}. The motivation is to make the input less accurate than $p^{(t)}$ by multiplying the initial estimate $p_\mathrm{sim}$ to it.

To compare $p^{(t)}$ and $p^{(t)}_{\mathrm{aug}}$ of Eq.~\ref{eq:dynamic_prior_3}, we conducted an experiment on PASCAL-5$^i$ 1-shot setting using ResNet-50; see the details of the experimental configuration for Section \ref{sec_exp}. Specifically, we run the network with  $p^{(0)}=p_\mathrm{sim}$ and then run it again with $p^{(1)}$ or with $p^{(1)}_\mathrm{aug}$. Table \ref{tab:ab_sim} shows the results, which includes the result of the first run, denoted by $p_\mathrm{sim}$; $p^{(1)}$ and $p^{(1)}_\mathrm{aug}$ indicate the results of the second run with them as the input. It is seen that the second run with $p^{(1)}$ does improve the estimation accuracy (i.e., from 60.8 to 61.8) and it is outperformed by $p^{(1)}_\mathrm{aug}$ (i.e., 62.7). From this result along with the results of our preliminary experiments, we choose $p^{(t)}_\mathrm{aug}$ in what follows. 


\begin{table}[t]
\caption{Accuracy of different inputs to the network. $p_\mathrm{sim}$ indicates the accuracy of the first run. $p^{(1)}$ and $p^{(1)}_\mathrm{aug}$ are those of the second run using them for the inputs.}
\centering
\begin{tabular}{c|ccc}
\hline\\[-2.0ex]
Estimate & $p_{\mathrm{sim}}$ & $p^{(1)}$ & 
$p^{(1)}_{\mathrm{aug}}$\\[0.5ex]
\hline
mIoU & 60.8 & 61.8 & 62.7 \\
\hline
\end{tabular}
\label{tab:ab_sim}
\end{table}




\subsubsection{Training Networks.}

The proposed method updates the estimate of the segmentation mask iteratively, starting from the initial guess of the similarity map. There are two choices in how to use the network $G$ at each iteration step, i.e., an identical network having the same weights or a network having the same architecture but different weights. The former simply runs the same network at every step. The latter runs different networks at the steps, which increases the representation power at the cost of memory. 

When the identical network $G$ with the weights $\theta$ is used at every step, we train it by 
\begin{equation}
\label{eq:dynamic_loss}
\min_{\theta} \frac{1}{T}
\sum_{t=1}^T \mathrm{CE}(p^{(t)}, y_{\mathrm{q}}), 
\end{equation}
where $\mathrm{CE}(\cdot)$ means the cross-entropy loss function and $y_\mathrm{q}$ is the ground-truth class (or the segmentaion mask).
When using a different network at each step, we train a network $G^{(t)}$ with weights $\theta^{(t)}$ at time $t$ sequentially. Specifically, we iterate the following minimization
\begin{equation}\label{eq:static_prior_1}
\min_{\theta^{(t)}} \mathrm{CE}(p^{(t)}, y_{\mathrm{q}})
\end{equation}
for $t=1,\ldots,T$.

\subsubsection{Settings for $K$-shot Segmentation.}

In the $K$-shot scenario, we are given $K$ support images. We calculate and use the average of the mid-level features from them and that of the similarity maps between each support image and the query. Specifically, for the query image $x_{\mathrm{q}}$ and every support sample $(x_{\mathrm{s}}^{(i)},y_{\mathrm{s}}^{(i)})$, we compute them as
\begin{equation}\label{eq:kshot}
\begin{aligned}
\Bar{f}_{\mathrm{ms}} &= \frac{1}{K}\sum\limits_{i=1}^{K}f_{\mathrm{ms}}^{(i)},\\
\Bar{p}_{\mathrm{sim}} &= \frac{1}{K}\sum\limits_{i=1}^{K}p_{\mathrm{sim}}^{(i)}.
\end{aligned}
\end{equation}
Then, we simply use $\Bar{f}_\mathrm{ms}$ and $\Bar{p}_\mathrm{sim}$ as ${f}_\mathrm{ms}$ and ${p}_\mathrm{sim}$, respectively. 

\section{Experiments}
\label{sec_exp}

\subsection{Experimental Settings}
\subsubsection{Datasets.}
We evaluate the performance of our method on PASCAL-5$^i$~\cite{bmvconeshot} and COCO-20$^i$~\cite{fwb}, which are standard datasets for few-shot segmentation.
PASCAL-5$^i$ includes images from PASCAL VOC 2012~\cite{pascalvoc} and additional annotations from SBD~\cite{sbd}. It consists of 20 categories and we divide them into 4 splits following the previous work~\cite{bmvconeshot,pfenet}.
The model evaluation is performed in a cross-validation fashion; three splits are used for training and the remaining one is used for evaluation.
In COCO-20$^i$, we divide the overall 80 categories in COCO~\cite{coco} into 4 splits and evaluate a model with the same cross-validation manner as for PASCAL-5$^i$.



\begin{table*}
\caption{{\color{black} Comparison with state-of-the-art methods under 1-shot and 5-shot settings on COCO-20$^i$. mIoU values for each test split and the averaged mIoU values (termed as mean) for four test splits are shown.}}
\centering
\begin{tabular}{l|c|cccc|c|cccc|c}
\hline
\multirow{2}{*}{Methods} & \multirow{2}{*}{BB.} & \multicolumn{5}{c}{1-shot} & \multicolumn{5}{|c}{5-shot} \\
\cline{3-12}
& & S0 & S1 & S2 & S3 & Mean & S0 & S1 & S2 & S3 & mean \\
\hline\hline
FWB~\cite{fwb} & VGG & 18.4 & 16.7 & 19.6 & 25.4 & 20.0 & 20.9 & 19.2 & 21.9 & 28.4 & 22.6 \\
PANet~\cite{panet} & VGG & - & - & - & - & 20.9 & - & - & - & - & 29.7 \\
PFENet~\cite{pfenet} & VGG & 33.4 & 36.0 & 34.1 & 32.8 & 34.1 & 39.2 & 47.1 & 41.5 & 40.4 & 42.1 \\
\rowcolor{lightgray}
Ours & VGG & \textbf{34.6} & \textbf{36.6} & \textbf{35.9} & \textbf{35.0} & \textbf{35.5} & \textbf{40.3} & \textbf{48.0} & \textbf{44.0} & \textbf{43.0} & \textbf{43.8} \\
\hline
RePRI~\cite{repri} & RES & 31.2 & 38.1 & 33.3 & 33.0 & 34.0 & 38.5 & 46.2 & 40.0 & 43.6 & 42.1 \\
ASGNet~\cite{asgnet} & RES & 34.9 & 36.9 & 34.3 & 32.1 & 34.6 & 41.0 & 48.3 & 40.1 & 40.5 & 42.5 \\
PFENet~\cite{pfenet} & RES & 35.7 & 41.4 & 38.9 & 35.4 & 37.9 & 38.6 & 47.7 & 45.2 & 40.3 & 43.0 \\
\rowcolor{lightgray}
Ours & RES & \textbf{37.5} & \textbf{41.4} & \textbf{40.0} & \textbf{38.1} & \textbf{39.3} & \textbf{42.2} & \textbf{49.9} & \textbf{47.3} & \textbf{46.3} & \textbf{46.4} \\
\hline
\end{tabular}
\label{tab:coco}
\end{table*}

\begin{table*}
\caption{{\color{black} Comparison with state-of-the-art methods under 1-shot and 5-shot settings on PASCAL-5$^{i}$. mIoU values for each test split and the averaged mIoU values (termed as mean) for four test splits are shown.}}
\centering
\begin{tabular}{l|c|cccc|c|cccc|c}
\hline
\multirow{2}{*}{Methods} & \multirow{2}{*}{BB.} & \multicolumn{5}{c}{1-shot} & \multicolumn{5}{|c}{5-shot} \\
\cline{3-12}
& & S0 & S1 & S2 & S3 & Mean & S0 & S1 & S2 & S3 & mean \\
\hline\hline
PANet~\cite{panet} & VGG & 42.3 & 58.0 & 51.1 & 41.2 & 48.1 & 51.8 & 64.6 & 59.8 & 46.5 & 55.7 \\
FWB~\cite{fwb} & VGG & 47.0 & 59.6 & 52.6 & 48.3 & 51.9 & 50.9 & 62.9 & 56.5 & 50.1 & 55.1 \\
RPMM~\cite{rpmm} & VGG & 47.1 & 65.8 & 50.6 & 48.5 & 53.0 & 50.0 & 66.5 & 51.9 & 47.6 & 54.0 \\
PFENet~\cite{pfenet} & VGG & 56.9 & 68.2 & 54.4 & 52.4 & 58.0 & 58.9 & 69.9 & 54.6 & 58.1 & 60.4 \\
\rowcolor{lightgray}
Ours & VGG & \textbf{57.7} & \textbf{70.1} & \textbf{56.0} & \textbf{56.0} & \textbf{60.0} & \textbf{60.5} & \textbf{71.3} & \textbf{56.8} & \textbf{60.4} & \textbf{62.3} \\
\hline
RePRI~\cite{repri} & RES & 60.2 & 67.0 & \textbf{61.7} & 47.5 & 59.1 & 64.5 & 70.8 & \textbf{71.7} & 60.3 & \textbf{66.8} \\
ASGNet~\cite{asgnet} & RES & 58.8 & 67.9 & 56.8 & 53.7 & 59.3 & 63.7 & 70.6 & 64.2 & 57.4 & 63.9 \\
PFENet~\cite{pfenet} & RES & 61.7 & 69.5 & 55.4 & 56.3 & 60.8 & 64.4 & 72.2 & 55.7 & 59.5 & 63.0 \\
SCL~\cite{scl} & RES & 63.0 & 70.0 & 56.5 & 57.7 & 61.8 & 64.5 & 70.9 & 57.3 & 58.7 & 62.9 \\
\rowcolor{lightgray}
Ours & RES & \textbf{63.7} & \textbf{70.4} & 57.3 & \textbf{59.2} & \textbf{62.7} & \textbf{66.2} & \textbf{72.6} & 58.7 & \textbf{63.0} & 65.1 \\
\hline
\end{tabular}
\label{tab:pascal}
\end{table*}

\subsubsection{Evaluation Metric.} 
Following the work~\cite{bmvconeshot}, we use mean intersection over union (mIoU) as an evaluation metric; mIoU is computed by taking average of the intersection-over-unions over different foreground objects in the test images.
We report the mIoU of each test split and averaged mIoU of four splits for comparison.

\subsection{Implementation Details}
We evaluate our methods on two backbone networks, VGG-16~\cite{vgg} and ResNet-50~\cite{resnet}, and they are pretrained on ImageNet~\cite{imagenet}. Our network is implemented by PyTorch~\cite{pytorch} and trained on a NVIDIA V100 GPU. 

We optimize our model using a SGD optimizer with momentum $=0.9$ and weight decay $=0.0001$.
For PASCAL-5$^i$, we train our model for 200 epochs with a learning rate of $0.0025$ and batch size $=4$. For COCO-20$^i$, we train the model for 50 epochs with a learning rate of $0.005$ and batch size $=8$. 
As a learning rate schedule, we adopt a polynomial decay with power $=0.9$ for both datasets.
As a data augmentation, we use random horizontal flip, random rotation, and then all images are randomly cropped to $473 \times 473$ pixels for PASCAL-5$^i$ and $641 \times 641$ pixels for COCO-20$^i$.



\subsection{Ablation Study}
To evaluate the effect of each component of our method, we perform the ablation study with a ResNet-50 backbone on PASCAL-5$^{i}$ and the 1-shot setting; the support set contains an image and its ground-truth mask.
\subsubsection{Effects of Multiple Refinements.}
As stated in Sec.~\ref{sec_method}, our method can refine the estimate multiple times if necessary. To explore the effect of the number of refinements $T$, we conduct experiments with the identical and the different network settings.
Table~\ref{tab:ab_iteration} shows that the performance improves as the number of refinements increases; the best results are achieved by $T=3$ for both identical and different network settings.
However, considering the performance improvement from $T=2$ to $T=3$ is slight and its computational cost, it is reasonable to perform the refinement with $T=2$. In the following all of our experiments, we show our results with $T=2$.

\begin{table}[H]
\caption{Effect of the number of refinements and network configurations. Numbers show the mIoU values on PALCAL-5$^i$ and the 1-shot setting.}
\centering
\begin{tabular}{c|c|c}
\hline
\# of refinements & Identical & Different \\
\hline
1 & 60.8 & 60.8 \\
2 & 61.4 & 62.7 \\
3 & 61.6 & 63.0 \\
\hline
\end{tabular}
\label{tab:ab_iteration}
\end{table}

\subsubsection{Identical Weights or Different Weights.}
We have two choices in how to use the network $G$ at each iteration step, i.e., a network having identical weights or a network having different weights but the same architecture.
As in Table~\ref{tab:ab_iteration}, the network with different weights outperforms the network having the identical weights by a good margin. Thus, we use the network $G$ having different weights for the following experiments.

\begin{figure*}
\includegraphics[width=1.0\textwidth]{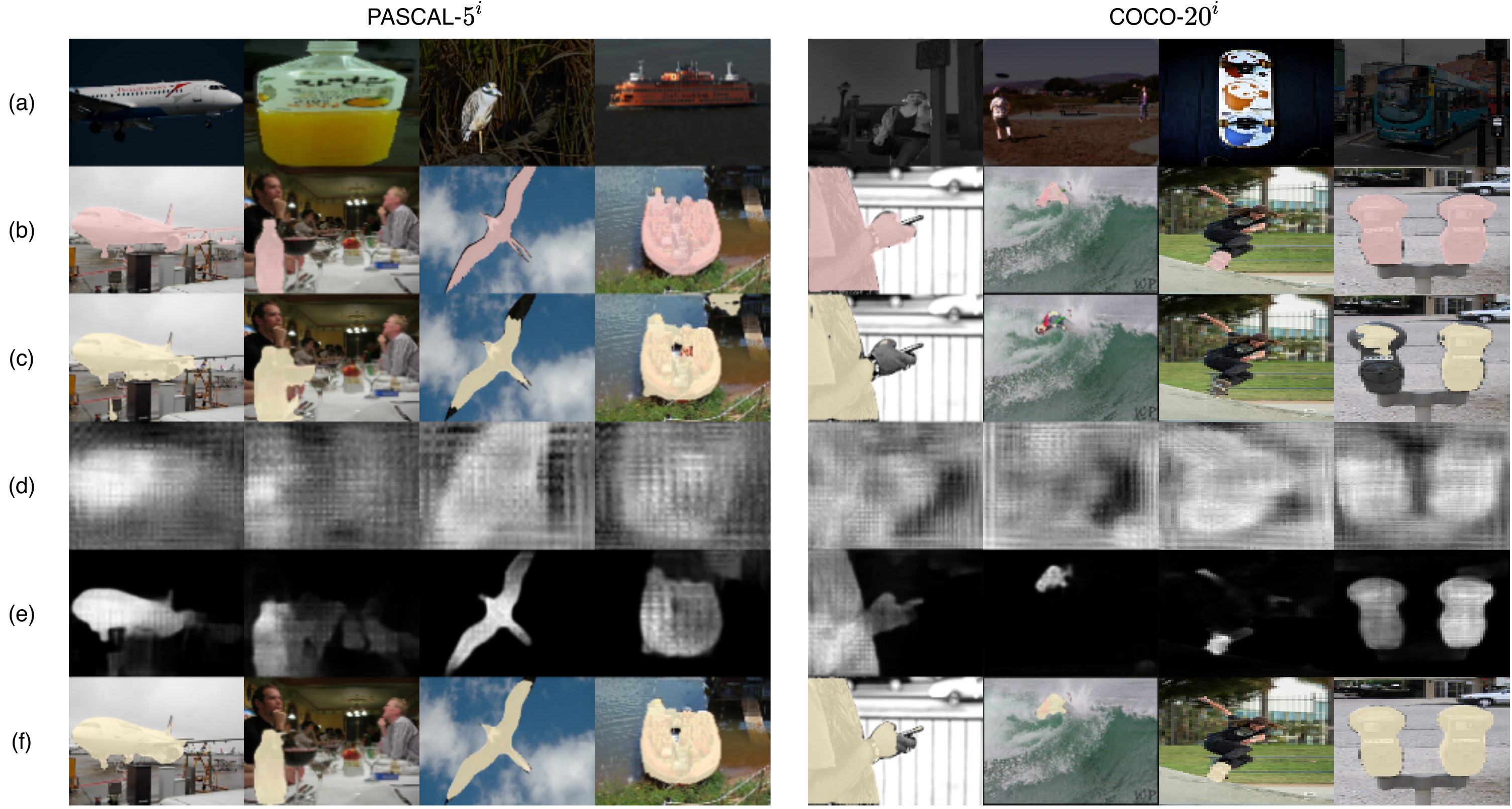} 
\caption{{\color{black} Visualization results of our method and PFENet in the 1-shot setting. The left panel is from PASCAL-5$^i$ and the right panel is from COCO-20$^i$. (a) Support images (foreground objects are highlighted). (b) Query images with their ground-truth masks. (c) PFENet results. (d) $p_{\mathrm{sim}}$ of Eq.~\ref{eq:cossim_cal_3}. (e) $p_{\mathrm{aug}}$ of Eq.~\ref{eq:dynamic_prior_3}. (f) Our method results.}}
\label{pic:vis}
\end{figure*}

{\color{black}
\subsection{Comparison with State-of-the-art Methods}
We compare our method with recently published methods on COCO-20$^i$ and PASCAL-5$^i$ datasets, including FWB~\cite{fwb}, PANet~\cite{panet}, PFENet~\cite{pfenet}, RePRI~\cite{repri}, ASGNet~\cite{asgnet}, and SCL~\cite{scl}, all of which are evaluated under the same evaluation metrics.

\subsubsection{COCO-20$^i$.} We report the average mIoU values that we calculated using randomly sampled $20,000$ episodes for each test split~\cite{pfenet}.
Table~\ref{tab:coco} shows the mIoU values with two backbone networks in the 1-shot and 5-shot settings. It can be seen that our approach yields the best performance in all scenarios and achieves new state-of-the-art performances on both 1-shot and 5-shot settings. Moreover, our approach significantly improves the performance of the baseline method (i.e., PFENet) on 5-shot setting from $43.0\%$ to $46.4\%$; we can also observe mIoU gains of $1.4\%$ in another setting. 

\subsubsection{PASCAL-5$^i$.} We report the average mIoU values that we calculated using randomly sampled $1,000$ episodes for each test split~\cite{bmvconeshot}.
As in Table~\ref{tab:pascal}, our method achieves the best performance on 1-shot setting and the second best on the 5-shot setting with the ResNet backbone.
Once again, our method yields significant improvement over VGG-16 based PFENet with mIoU increases of around $2\%$ in all settings.


\subsubsection{Qualitative Results.}
We randomly select several examples from the test episodes in PASCAL-5$^i$ and COCO-20$^i$, and visualize the segmentation results under the 1-shot setting. Fig.~\ref{pic:vis} shows the qualitative comparisons of the proposed method and PFENet along with $p_{sim}$ and $p_{aug}$. It is seen that our method can generate accurate segmentation masks regardless of the object size or appearance. In particular, when focusing on small objects (e.g., person and skateboard in the right panel in Fig.~\ref{pic:vis}), we can observe that our method predicts much more accurate masks than those by PFENet. As seen in Fig.~\ref{pic:vis} (d) and (e), the similarity maps $p_{sim}$ used in PFENet do not precisely focus on the target objects due to their roughness while our approach does. This agrees with the improvement on the mIoU values shown in Table~\ref{tab:coco} and \ref{tab:pascal}.
}

\section{Summary and Conclusion}

In this paper, we have considered few-shot segmentation. The key to its success is how to use the mid-level and high-level features extracted by a pretrained CNN from the input query and support images. Recent few-shot segmentation methods employ the approach proposed by Tsai et al., which is to infer the object region using the mid-level features of the input images under the guidance from the similarity map of the high-level features between the query and support images. In this paper, we have shown a novel interpretation of the approach. Specifically, we regard the similarity map computed from the high-level features as a rough, initial estimate of the object region. We then regard that the approach improves the estimate (especially in its inaccurate segmentation boundaries) with the help of the mid-level features. In this interpretation, the high-level features play a leading role, and the mid-level features are given a supporting role, which is the opposite of the previous explanation. Based on this reinterpretation, we propose a method for improving estimation accuracy simply by repeatedly applying the network multiple times to the latest estimate, starting from an initial estimate. We have experimentally obtained several observations. First, it yields better results when we input not the latest estimate of the object region but its multiplication with the initial estimate to the network. Second, the accuracy improvement by the repeated network application is mostly saturated with a few repetition counts. Third, although we use the network(s) with the same architecture for all the iterative steps, using networks with different weights yields better results than using an identical network with the same weights at different steps. Finally, we have shown that the proposed approach achieves the new state-of-the-art on multiple standard benchmark tests of few-shot segmentation.

\bibliography{ref.bib}

\end{document}